\documentclass{article}
\usepackage{arxiv}
\usepackage{graphicx}
\usepackage{mathptmx}
\usepackage[subrefformat=parens,labelformat=parens]{subcaption}
\usepackage{authblk}
\usepackage{amsmath}
\usepackage{amsfonts}
\usepackage{amssymb}
\usepackage{amsbsy}
\usepackage{amsthm}
\usepackage{booktabs}
\usepackage{comment}
\usepackage{multirow}
\usepackage[ruled,vlined]{algorithm2e}
\usepackage{bbm}
\usepackage{wrapfig}
\usepackage{lipsum}
\usepackage{hyperref}
\allowdisplaybreaks

\newtheorem{theorem}{Theorem}

\newtheorem{remark}[theorem]{Remark}

\begin{document}

\title{\Large
Linear Noise Approximation Assisted Bayesian Inference on Mechanistic Model of Partially Observed Stochastic Reaction Network}

\author[1]{Wandi Xu}
\author[  ~,1]{Wei Xie\thanks{Corresponding author: w.xie@northeastern.edu}}
\affil[1]{Northeastern University, Boston, MA 02115}

\maketitle

\begin{abstract}
To support mechanism online learning and facilitate digital twin development for biomanufacturing processes, this paper develops an efficient Bayesian inference approach for partially observed enzymatic stochastic reaction network (SRN), a fundamental building block of multi-scale bioprocess mechanistic model. To tackle the critical challenges brought by the nonlinear stochastic differential equations (SDEs)-based mechanistic model with partially observed state and having measurement errors, an interpretable Bayesian updating linear noise approximation (LNA) metamodel, incorporating the structure information of the mechanistic model, is proposed to approximate the likelihood of observations. Then, an efficient posterior sampling approach is developed by utilizing the gradients of the derived likelihood to speed up the convergence of Markov chain Monte Carlo (MCMC). The empirical study demonstrates that the proposed approach has a promising performance.
\end{abstract}


\section{INTRODUCTION}
\label{sec:intro}

Partially observed stochastic reaction network (SRN) modeling the dynamics of a population of interacting species, such as chemical molecules participating in multiple reactions, is the fundamental building block of \textit{multi-scale bioprocess mechanistic model} characterizing the causal interdependences from molecular- to macro-kinetics. It plays a critical role to: (1) facilitate digital twin development and support mechanism learning for biomanufacturing processes; (2) allow us to probe critical latent state based on partially observed information; and (3) serve as a fundamental model for a biofoundry platform \cite{hillson2019building} that can integrate heterogeneous online and offline measures collected from different manufacturing processes and speed up the bioprocess development with much less experiments. Model inference on the SRN mechanistic model based on heterogeneous data also helps to strengthen the theoretical foundations of federated learning on bioprocess mechanisms, through which we can train and advance knowledge.

The SRN mechanistic model has three key features that make the model inference challenging. First, the continuous-time state transition model, representing the evolution of concentration or number of molecules, is highly nonlinear. At any time, the reaction rates, characterizing the regulation mechanisms of enzymatic reaction network, are a function of random state. We adopt the diffusion approximation in \cite{gillespie2000chemical} to model the state dynamics with a set of coupled stochastic differential equations (SDEs). In this case, the state transition model has double-stochasticity, making it analytically intractable to obtain the state transition densities at different times and also hard to get the closed form likelihood of observations. Second, since the state is partially observed, we need to integrate out the unobserved state variables to get the likelihood. Third, the data collected from biomanufacturing processes are heterogeneous and also subject to measurement errors.

The model inference of enzymatic SRN has found increasing interest especially in biomanufacturing digital twin development. Even under the situations with the reaction network structure known, that is built on thousand years of the understanding on biological system mechanisms, the mechanistic model parameters are often unknown. It is necessary to infer these parameters using the observations collected from biomanufacturing processes. Since each batch of production can be expensive, we often have very small amount of experimental observations. Coupled with high stochasticity of biomanufacturing processes, the model uncertainty tends to be high. However, frequentist model uncertainty quantification approaches are built on asymptotic approximation, such as asymptotic normality and bootstrap. Thus, in this paper, we focus on a Bayesian inference on multi-scale mechanistic model, which can support online learning and interpretability.

An enormous volume of literature has been dedicated to Bayesian inference for SRN mechanistic model. As the exact state transition density of the SDEs-based mechanistic model is unknown, coupled with another challenge (i.e., the partially observed state), the marginal likelihood integrating out the unobserved state variables is intractable. Thus, many existing works are sampling approaches without the explicit calculations of the likelihood, such as approximate Bayesian computation (ABC) and its variants \cite{xie2022sequential}. But in sampling approaches without using likelihood, the complex structure and high stochasticity of SRN make the simulation generating a large amount of sample paths computationally expensive and the acceptance rates of samplers very low. \textit{Therefore, we construct a metamodel to approximate the state transition densities of the SDEs, obtain a likelihood approximation, and utilize it to speed up Bayesian inference.}

Gaussian Process (GP) is often used as a metamodel. The studies \cite{archambeau2007gaussian} and \cite{garcia2017nonparametric} use GPs as priors for nonparametric estimation of the drift and diffusion terms of SDEs without an exact knowledge of their functional forms. In this paper, we suppose the structure of SDEs-based mechanistic model is known. To completely exploit such structure information and improve the interpretability of constructed metamodel, we refer to the deterministic ordinary differential equation (ODE)-based dynamic system inference. In particular, \cite{yang2021inference} specifies a GP prior over the solution to the ODE, and restricts the GP on a manifold that satisfies the ODE system, to address the incompatibility between the metamodel and the mechanistic model. And an alternative to GP under SDE-based model is linear noise approximation (LNA). The LNA was originally proposed to approximate the solution of the chemical master equation (CME) \cite{gillespie1992rigorous}, and it can be derived in a number of ways. For instance, \cite{ferm2008hierarchy} and \cite{ruttor2009efficient} follow the idea of an asymptotic system size expansion, and derive the LNA by approximating the CME through a Taylor expansion. Since the solution of SDE itself is a random variable, it is difficult to extend the GP approach developed in \cite{yang2021inference} to the SDE model inference. Instead, following \cite{fearnhead2014inference}, we specify the derived LNA as a prior to the solution of the SDE, through which we take full advantage of the structure information provided by the SDE model without the time-consuming numerical integration.

The likelihood of observations can be obtained under the LNA, but the exact Bayesian posterior is still not analytically tractable as a conjugate prior is hard to find. One thus defers to sampling approaches to generate samples from the posterior. The most common one is Markov chain Monte Carlo (MCMC), such as Metropolis-Hastings algorithm. Its effectiveness depends heavily on the choice of the proposal distribution. Metropolis-adjusted Langevin algorithm (MALA) makes use of the additional gradient information of the target posterior distribution to construct a better proposal distribution, which is shown to have a faster mixing time compared with classic MCMC \cite{chewi2021optimal}. Therefore, in this paper, we specifically tailor a MALA procedure to generate posterior samples more efficiently.

In specific, we propose a LNA assisted Bayesian inference on the nonlinear multivariate SDE-based mechanistic model with partially observed state and subject to measurement errors. The main contributions are twofold. First, an interpretable Bayesian updating LNA metamodel is developed for likelihood approximation. It provides a coherent way to simultaneously satisfy the SDE model and fit the observed data, allowing us to probe critical latent state based on partially observed information. Second, the proposed MALA procedure utilizes the gradient information from the derived likelihood to speed up MCMC search and more efficiently generate posterior samples. The proposed Bayesian inference for SRN can support online mechanism learning, facilitate digital twin development, and speed up bioprocess design and control.

The paper is organized as follows. We provide a brief introduction of the SDE-based mechanistic model for enzymatic SRN and problem description in Section~\ref{sec:problemDescription}. To facilitate the model Bayesian inference, the LNA is used to construct the state transition densities and a closed form likelihood is thus derived in Section~\ref{sec:linearNoiseApprox}. Then, we propose an efficient and interpretable Bayesian posterior sampling algorithm in Section~\ref{sec:BayesianInference}. Its performance is studied in Section~\ref{sec:empirical}. Finally, we conclude the paper in Section~\ref{sec: conclusion}.

\section{STOCHASTIC REACTION NETWORK (SRN) MODEL AND PROBLEM DESCRIPTION}
\label{sec:problemDescription}

\textbf{(1) SRN model.}
We first review a general SRN composed of $J$ species, denoted by $\pmb{X}=(X_1, X_2, \ldots, X_{J})^\top$, interacting with each other through $K$ reactions. The number of molecules of species $j$ at time $t$ is denoted by $x_j(t)$ and $\pmb{x}(t)=\left(x_1(t),x_2(t),\ldots,x_J(t)\right)^\top$. Each reaction is characterized by a nonzero reaction vector $\pmb{C}_k \in \mathbb{R}^{J}$ for $k=1,2\ldots,K$, describing the change in the numbers of $J$ species' molecules when a $k$-th molecular reaction occurs. The associated propensity function, denoted by $\omega_k$, describes the probability with which the $k$-th reaction occurs per time unit. Specifically, for the $k$-th reaction equation given by
\begin{equation*}
    p_{k1} X_1 + p_{k2} X_2 + \cdots + p_{kJ} X_{J} \xrightarrow{\omega_k} q_{k1} X_1 + q_{k2} X_2 + \cdots + q_{kJ} X_{J},
\end{equation*}
the reaction relational structure, specified by $\pmb{C}_k = (q_{k1}-p_{k1},q_{k2}-p_{k2},\ldots,q_{kJ}-p_{kJ})^\top$, is known for $k=1,2,\ldots,K$. Thus, the \textit{stoichiometry matrix} $\pmb{C} = \left(\pmb{C}_1, \pmb{C}_2, \ldots, \pmb{C}_{K}\right) \in \mathbb{R}^{J \times K}$ characterizes the structure information of the reaction network composed of $K$ reactions, where its $(i,j)$-th element represents the number of molecules of the $i$-th species that are either consumed (indicated by a negative value) or produced (indicated by a positive value) in each random occurrence of the $j$-th reaction.

Then, we describe the state transition model for bioprocess. As a multi-scale bioprocess representing the dependence from molecular- to macro-kinetics, it is built on the fundamental building block, i.e., molecular reaction network. Let $d\pmb{R}(t)$ represent a $K$-dimensional vector of occurrences of each molecular reaction in an infinitesimal time interval $(t, t + dt]$. It follows a distribution with parameters depending on the propensity functions $\pmb{\omega}(\pmb{x}(t); \pmb{\theta}) = \left( \omega_1(\pmb{x}(t);\pmb{\theta}_1), \omega_2(\pmb{x}(t);\pmb{\theta}_2), \ldots, \omega_{K}(\pmb{x}(t);\pmb{\theta}_{K}) \right)^\top$, where the structure of each $\omega_k(\pmb{x}(t);\pmb{\theta}_k)$, characterizing the bioprocess regulation mechanism for the $k$-th molecular reaction, is given and we focus on the inference of the unknown parameters $\pmb{\theta} = (\pmb{\theta}_1^\top,\pmb{\theta}_2^\top,\ldots, \pmb{\theta}_{K}^\top)^\top$. 

Due to the fact that reaction events change species numbers by an integer amount, the state transition model is naturally characterized by a continuous-time Markov jump process \cite{anderson2011continuous}. In particular, assuming that two reactions cannot occur at exactly the same time, one can represent the occurrences number of each $k$-th reaction in an infinitesimal time interval $(t, t + dt]$, denoted by $dR_k(t)$ (i.e., the $k$-th component of $d\pmb{R}(t)$), using one of the most elementary counting process, namely, the nonhomogeneous Poisson process. Since the dynamic change of propensity function in any infinitesimal time interval $(t, t + dt]$ is negligible, the intensity of $dR_k(t)$ becomes $\omega_k (\pmb{x}(t);\pmb{\theta}_k) dt$. And conditional on $\pmb{x}(t)$, $dR_k(t)$ for $k=1,2,\ldots,K$ can be considered as independent of one another and are Poisson$(\omega_k (\pmb{x}(t);\pmb{\theta}_k) dt)$ random variables, from which we have $\mathbb{E}(d\pmb{R}(t)) = \pmb{\omega} (\pmb{x}(t); \pmb{\theta}) dt$ and Cov$(d\pmb{R}(t)) = {\rm diag}\{\pmb{\omega} (\pmb{x}(t); \pmb{\theta})\}dt$. Under the Poisson assumption, we adopt the \textit{diffusion approximation to Markov jump process} following the study \cite{gillespie2000chemical} and then model $d\pmb{R}(t)$ with It\^{o} SDE, i.e., 
\begin{equation*}
    d\pmb{R}(t) = \mathbb{E}(d\pmb{R}(t)) + \left\{{\rm Cov}(d\pmb{R}(t))\right\}^{\frac{1}{2}}d\pmb{B}(t) = \pmb{\omega} (\pmb{x}(t); \pmb{\theta}) dt + \left\{{\rm diag}\{\pmb{\omega} (\pmb{x}(t); \pmb{\theta})\}\right\}^{\frac{1}{2}}d\pmb{B}(t),
\end{equation*}
where $d\pmb{B}(t)$ is the increment of a $K$-dimensional standard Brownian motion. Given the reaction network structure specified by the stoichiometry matrix $\pmb{C}$, the impact on the process dynamics becomes, 
\begin{equation}
    d\pmb{x}(t) = \pmb{C}d\pmb{R}(t) = \pmb{C} \pmb{\omega}(\pmb{x}(t); \pmb{\theta}) dt + \left\{\pmb{C}{\rm diag}\{\pmb{\omega} (\pmb{x}(t); \pmb{\theta})\}\pmb{C}^\top\right\}^{\frac{1}{2}} d\pmb{B}(t). \label{diffusion approximation - x}
\end{equation}

For both theoretical study and practical application purposes, the system is assumed to have a size parameter $\Omega$ (such as the volume of bioreactor). Then $s_j(t) = x_j(t)/\Omega$ represents the concentration of molecules of species $j$. At any time $t$, let $\pmb{s}(t) = (s_1(t),s_2(t),\ldots,s_{J}(t))^\top = \Omega^{-1}\pmb{x}(t)$ be the bioprocess state. And the propensity functions $\omega_k(\pmb{x}(t);\pmb{\theta}_k)$ for $k=1,2,\ldots,K$ can be written as
\begin{equation}
    \omega_k(\pmb{x}(t);\pmb{\theta}_k) = \Omega v_k\left(\Omega^{-1}\pmb{x}(t);\pmb{\theta}_k\right) = \Omega v_k(\pmb{s}(t);\pmb{\theta}_k), \label{propensity_rate}
\end{equation}
where $v_k$ is the reaction rate associated with the $k$-th reaction, specified by the parameters $\pmb{\theta}_k$ and depending on the current system state $\pmb{s}(t)$. By plugging the relation between the propensity function and the reaction rate (i.e., Equation \eqref{propensity_rate}) into Equation \eqref{diffusion approximation - x}, we get the state transition,
\begin{align}
    d\pmb{s}(t) = \Omega^{-1} d\pmb{x}(t) &= \pmb{C} \pmb{v}(\pmb{s}(t); \pmb{\theta}) dt + \Omega^{-\frac{1}{2}} \left\{\pmb{C}{\rm diag}\{\pmb{v} (\pmb{s}(t); \pmb{\theta})\}\pmb{C}^\top\right\}^{\frac{1}{2}} d\pmb{B}(t) \nonumber \\
    &\triangleq \pmb{\mu}(\pmb{s}(t);\pmb{\theta}) dt + \Omega^{-\frac{1}{2}} \left\{\pmb{D}(\pmb{s}(t);\pmb{\theta})\right\}^{\frac{1}{2}} d\pmb{B}(t), \label{diffusion approximation}
\end{align}
where $\pmb{v}(\pmb{s}(t); \pmb{\theta}) = \left( v_1(\pmb{s}(t);\pmb{\theta}_1), v_2(\pmb{s}(t);\pmb{\theta}_2), \ldots, v_{K}(\pmb{s}(t);\pmb{\theta}_{K}) \right)^\top$ is the reaction rate vector. Equation \eqref{diffusion approximation} also represents the \textit{doubly stochastic property} of SRN, that is, both mean $\pmb{\mu}(\pmb{s}(t); \pmb{\theta})$ and variance $\pmb{D}(\pmb{s}(t); \pmb{\theta})$ are functions of the current system state $\pmb{s}(t)$ and characterized by the parameters $\pmb{\theta}$, while $\pmb{s}(t)$ is a random state vector that changes over time and its evolution (i.e., $d\pmb{s}(t)$) is characterized by $\pmb{\mu}(\pmb{s}(t);\pmb{\theta})$ and $\pmb{D}(\pmb{s}(t);\pmb{\theta})$.

\begin{figure}[htbp]
    \centering
    \subfloat[Sample observations]{\includegraphics[width=0.31\columnwidth]{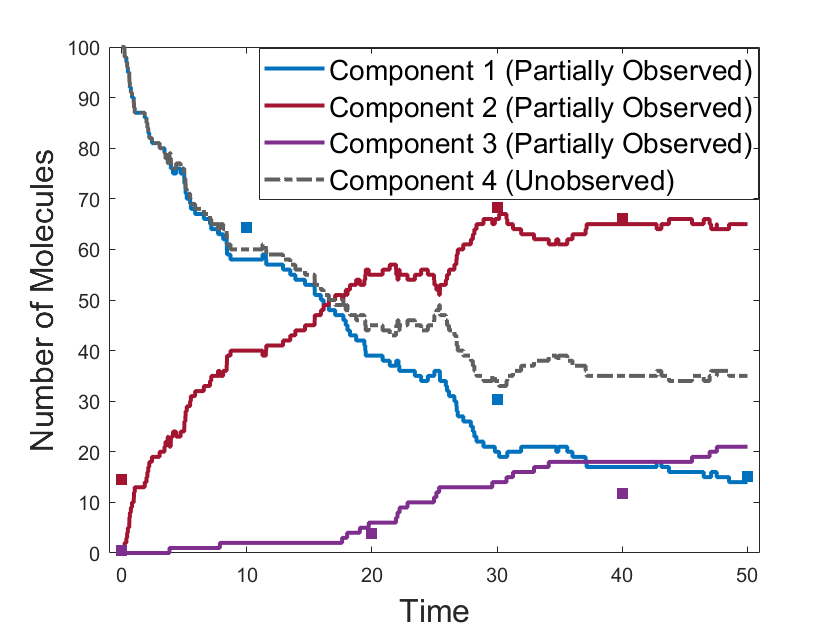}}
    \subfloat[Bayesian updating linear noise approximation (LNA) metamodel]{\includegraphics[width=0.69\columnwidth]{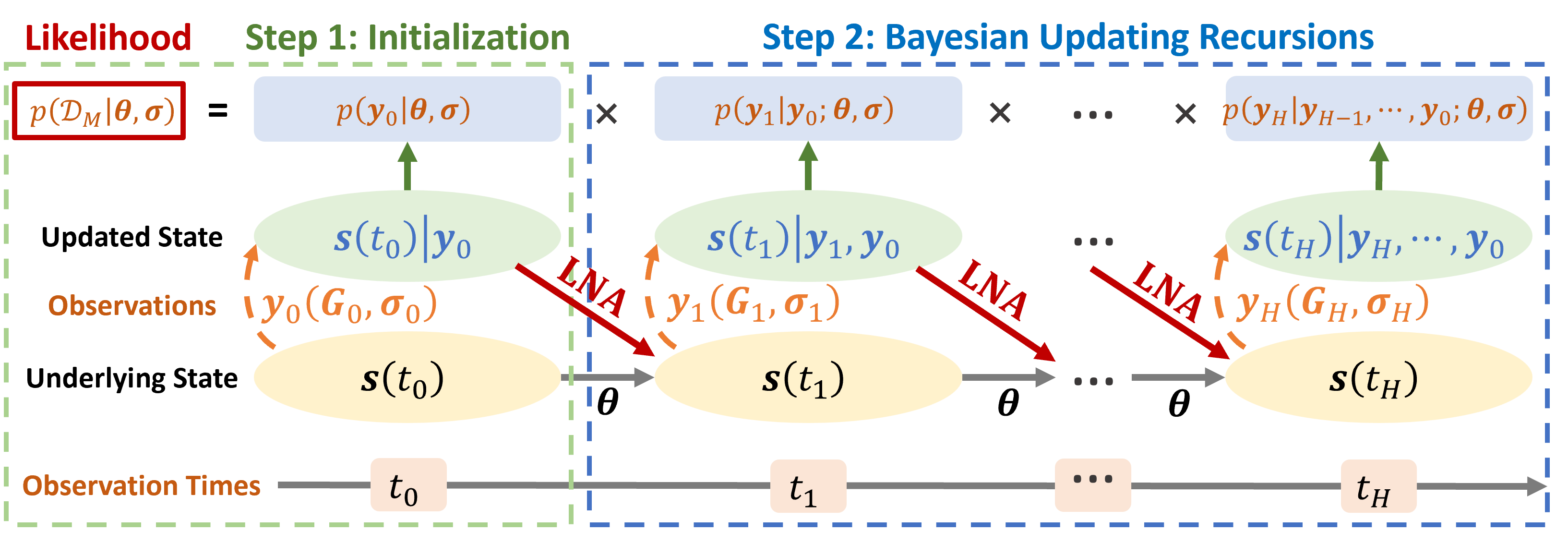}}
    \caption{An illustration of (a) the partially observed state with measurement error; and (b) the proposed interpretable Bayesian updating LNA metamodel for enzymatic stochastic reaction network (SRN).}
    \label{fig:intro}
\end{figure}

\vspace{0.05in}
\noindent \textbf{(2) Partially observed state and heterogeneous data collection.}
The measures of partially observed state variables are often heterogeneous and subject to measurement errors. The observations for different observable state components are also asynchronous; see Figure \ref{fig:intro}(a). In particular, we represent all observation times of state as the time set, denoted by $\pmb{T} = \{t_0,t_1,\ldots,t_H\}$, where $t_0 < t_1 < \cdots < t_H$, and the time intervals $\Delta t_h = t_{h+1}-t_h$ can be variable for $h=0,1,\ldots,H-1$. At each observation time $t_h$, we denote the set of observed components' subscripts of underlying state $\pmb{s}$ by $\pmb{J}_h$, i.e., $\pmb{J}_h = \{j \in [J]: s_j \text{ is observed at time } t_h\}$ where $[J]$ represents $\{1,2,\ldots,J\}$, and let $\pmb{J}_y = \cup_{h=0}^H \pmb{J}_h$ be the set of subscripts of the components that can be observed at certain times of experiments. The observations are denoted by $\pmb{y}_h(t_h) \in \mathbb{R}^{M|\pmb{J}_h|}$, where $|\pmb{J}_h| \leq J$ is the cardinality of $\pmb{J}_h$ representing the dimension of observed components of underlying state $\pmb{s}$ at time $t_h$, and $M$ is the batch size of experiments. Then, the observations at time $t_h$ can be modeled as 
\begin{equation}
    \pmb{y}_h(t_h) = \pmb{G}_h \pmb{s}(t_h) + \pmb{\epsilon}_h(t_h). \label{measurement error}
\end{equation}

Suppose the measurement errors follow a multivariate Gaussian distribution $\pmb{\epsilon}_h(t_h) \sim \mathcal{N}(\pmb{0}, \pmb{\Sigma}_h)$, where $\pmb{\Sigma}_h$ is a diagonal matrix with $M$ vectors of $\pmb{\sigma}_h$ on the main diagonal, and $\pmb{\sigma}_h = \left\{\sigma_{jj}: j \in \pmb{J}_h\right\}^\top$ is the vector of measurement error level at time $t_h$. Further, let $\pmb{\sigma} = \left\{\sigma_{jj}: j \in \pmb{J}_y\right\}^\top$ be the vector of measurement error level of all observed components. And $\pmb{G}_h$ is a $M|\pmb{J}_h|$-by-$J$ constant matrix, mapping the entire $J$-dimensional vector of underlying state $\pmb{s}(t_h)$ into the $M$ batches of $|\pmb{J}_h|$-dimensional vector containing only the counterpart of observed components at time $t_h$. Notice that the dimension $|\pmb{J}_h|$ can change at different observation times accounting for the fact that the measures of partially observed state are asynchronous.

Given the observed data set denoted by $\mathcal{D}_M = \{\pmb{y}_h(t_h)\}_{h=0}^{H}$, the model uncertainty is quantified by a posterior distribution $p\left(\pmb{\theta},\pmb{\sigma} | \mathcal{D}_M\right) \propto  p\left(\pmb{\theta}\right) p\left(\pmb{\sigma}\right) p\left(\mathcal{D}_M | \pmb{\theta},\pmb{\sigma}\right)$. With the collection of new experiment data $\Delta \mathcal{D}$, the model uncertainty can be updated as $p\left(\pmb{\theta},\pmb{\sigma} | \mathcal{D}_M \cup \Delta \mathcal{D}\right) \propto  p\left(\pmb{\theta},\pmb{\sigma} | \mathcal{D}_M\right) p\left(\Delta \mathcal{D} | \pmb{\theta},\pmb{\sigma}\right)$.

\vspace{0.05in}
\noindent \textbf{(3) Key challenges on Bayesian inference and summary of the proposed inference approach.} Our focus in this paper is to develop a computationally efficient Bayesian inference approach on unknown model parameters $\pmb{\theta} \in \pmb{\Theta}^{N}$ with emphasis on nonlinear $\pmb{\mu}(\pmb{s}(t); \pmb{\theta})$ and $\pmb{D}(\pmb{s}(t); \pmb{\theta})$ characterizing the regulation mechanisms of SRN as shown in \eqref{diffusion approximation}, where $\pmb{\Theta}^{N} \subset \mathbb{R}^N$ is the feasible parameter space. The first challenge is partially observed state subject to random measurement error. The observed components of system state $\pmb{s}$ are recorded at limited discrete time points and the observation time points of each observable component may not be synchronized; see Figure \ref{fig:intro}(a). Moreover, there are often some components of state $\pmb{s}$ unobservable. To tackle this challenge, we develop an interpretable \textit{Bayesian updating LNA metamodel} on underlying state $\pmb{s}(t)$ in Section~\ref{sec:linearNoiseApprox} so that we can predict all components of $\pmb{s}(t)$ at any time $t$. 

Such a metamodel needs to have capability to characterize the dependence between components of $\pmb{s}(t)$ and handle the doubly stochasticity of SRN. Luckily, we have a SDE model \eqref{diffusion approximation} representing the mechanism of state change. This brings us to the second challenge. On the one hand, the nonlinear drift and diffusion terms of the SDE \eqref{diffusion approximation} make solving it directly to get the metamodel of $\pmb{s}(t)$ require time-consuming numerical integration methods. On the other hand, the regulation mechanism structure information from the SDE cannot be completely exploited which impacts on interpretability if we choose a black-box metamodel. To take full advantage of the structure information about the state transition provided by the SDE \eqref{diffusion approximation}, and to avoid the use of numerical integration to solve these SDEs, we place LNA priors on the dynamics of state $\pmb{s}(t)$ to facilitate inference of model parameters $\pmb{\theta}$. Under the LNA, the underlying process $\{\pmb{s}(t): t\geq 0\}$ follows a multivariate Gaussian distribution, combined with the assumption of linear Gaussian relation between each observation $\pmb{y}_h(t_h)$ and underlying state value $\pmb{s}(t_h)$ as shown in Equation \eqref{measurement error} to give a tractable approximation to the likelihood of the observations $\{\pmb{y}_h(t_h)\}_{h=0}^{H}$. The key to our approach is, to avoid a poor approximation to the true distribution of $\pmb{s}(t)$ as $t$ gets large, we reinitialize the LNA for each time interval $(t_h,t_{h+1}]$ using the derived posterior distribution of $\pmb{s}(t_h)$ given $\pmb{y}_h(t_h),\pmb{y}_{h-1}(t_{h-1}),\ldots,\pmb{y}_0(t_0)$; see Figure \ref{fig:intro}(b).

\section{BAYESIAN UPDATING LINEAR NOISE APPROXIMATION (LNA) METAMODEL}
\label{sec:linearNoiseApprox}

In this section, we first utilize the LNA to approximate the SDE model \eqref{diffusion approximation} and then develop a Bayesian updating LNA metamodel to reduce the approximation error between the true solution to the SDE \eqref{diffusion approximation} and LNA model. The LNA divides the path $\{\pmb{s}(t): t \geq 0\}$ of the SDE \eqref{diffusion approximation} into a deterministic path $\{\Bar{\pmb{s}}(t): t \geq 0\}$ and a stochastic perturbation $\{\pmb{\xi}(t): t \geq 0\}$, where the fluctuations in $\pmb{s}(t)$ at any given time $t$ are assumed to be of $O(\Omega^{-\frac{1}{2}})$; see \cite{ferm2008hierarchy} and \cite{fearnhead2014inference} for a rigorous derivation and detailed discussion. Under this partition, through a Taylor expansion of the SDE \eqref{diffusion approximation} around $\Bar{\pmb{s}}(t)$ up to order $\Omega^{-\frac{1}{2}}$, we split the SDE \eqref{diffusion approximation} into one deterministic ODE with the solution $\Bar{\pmb{s}}(t)$ as shown in Equation \eqref{ODE},
\begin{equation}
    d \Bar{\pmb{s}}(t) = \pmb{\mu}(\Bar{\pmb{s}}(t);\pmb{\theta}) dt \label{ODE}
\end{equation}
with initial value $\Bar{\pmb{s}}(0)$, and one SDE with its solution $\pmb{\xi}(t)$ following a Gaussian distribution for any fixed or Gaussian distributed initial condition on $\pmb{\xi}(0)$, denoting by $\pmb{\xi}(t) \sim \mathcal{N} \left(\pmb{\varphi}(t), \pmb{\Psi}(t)\right)$. And its mean vector $\pmb{\varphi}(t)$ and covariance matrix $\pmb{\Psi}(t)$ for any $t \geq 0$ can be obtained by solving the ODEs in \eqref{SDE_mean} and \eqref{SDE_cov},
\begin{align}
    d\pmb{\varphi}(t) &= \nabla_{\pmb{s}} \pmb{\mu}(\pmb{s};\pmb{\theta})|_{\pmb{s}=\Bar{\pmb{s}}(t)} \pmb{\varphi}(t) dt, \label{SDE_mean} \\
    d\pmb{\Psi}(t) &= \left\{ \pmb{\Psi}(t) \left(\nabla_{\pmb{s}} \pmb{\mu}(\pmb{s};\pmb{\theta})|_{\pmb{s}=\Bar{\pmb{s}}(t)}\right)^\top + \nabla_{\pmb{s}} \pmb{\mu}(\pmb{s};\pmb{\theta})|_{\pmb{s}=\Bar{\pmb{s}}(t)} \pmb{\Psi}(t) + \pmb{D}(\Bar{\pmb{s}}(t);\pmb{\theta}) \right\} dt, \label{SDE_cov}
\end{align}
with initial values $\pmb{\varphi}(0)$ and $\pmb{\Psi}(0)$. Without loss of generality, in the following discussion, we simplify the notation and assume an unit system size $\Omega = 1$. Suppose the initial condition for the SDE \eqref{diffusion approximation} with $\Omega = 1$ is $\pmb{s}(0) \sim \mathcal{N} \left(\pmb{\alpha}^*(0), \pmb{\beta}^*(0)\right)$, then for arbitrary $\Bar{\pmb{s}}(0)$, we can set $\pmb{\varphi}(0)=\pmb{\alpha}^*(0)-\Bar{\pmb{s}}(0)$ and $\pmb{\Psi}(0)=\pmb{\beta}^*(0)$. Integrating the ODEs \eqref{ODE}, \eqref{SDE_mean}, and \eqref{SDE_cov} through time 0 to $t$ provides the LNA
\begin{equation}
    \pmb{s}(t) \sim \mathcal{N} \left(\Bar{\pmb{s}}(t) + \pmb{\varphi}(t), \pmb{\Psi}(t)\right). \label{dist_LNA}
\end{equation}

Under the LNA model \eqref{dist_LNA} on the partially observed state $\pmb{s}(t_h)$ with measurement error $\pmb{\epsilon}_h(t_h)$ as shown in \eqref{measurement error} for $h=0,1,\ldots,H$, the likelihood of the observations $\mathcal{D}_M$ is tractable. In particular, the ODE components of the LNA (i.e., Equations \eqref{ODE}, \eqref{SDE_mean}, and \eqref{SDE_cov}) are solved once over the entire time interval for given initial values. However, LNA can lead to a poor approximation to the true $\pmb{s}(t)$, due to the approximation error between the true solution to the SDE \eqref{diffusion approximation} and the LNA \eqref{dist_LNA} gradually accumulates as $t$ gets large.

To tackle this issue, we construct the likelihood of the observations $\mathcal{D}_M$ through using the updated LNA model at each observation time point $t_h$ with $h=0,1,\ldots,H$. In particular, given an estimate of model parameters $\pmb{\theta}$ and measure error level $\pmb{\sigma}$, we first set the LNA model \eqref{dist_LNA} with the initial condition $\pmb{s}(t_0) \sim \mathcal{N} \left(\Bar{\pmb{s}}(t_0)+\pmb{\varphi}(t_0), \pmb{\Psi}(t_0)\right)$ as a prior, and then the observations $\pmb{y}_h(t_h) \in \mathcal{D}_M$ are used to \textit{sequentially} update the prior on $\pmb{s}(t_h)$ for each $t_h$ with the procedure shown in Figure \ref{fig:intro}(b). Therefore, we can approximate the distribution of $\pmb{y}_h(t_h)$ given all observations up to time $t_h$ and obtain the likelihood. The detailed procedure is summarized in the following three steps.

\textbf{Step 1:} At the initial observation time point $t_0$, given the prior $\pmb{s}(t_0) \sim \mathcal{N} \left(\Bar{\pmb{s}}(t_0)+\pmb{\varphi}(t_0), \pmb{\Psi}(t_0)\right)$ and the observational uncertainty \eqref{measurement error}, we can directly have
\begin{equation}
    \pmb{y}_0(t_0) | \pmb{\sigma} \sim \mathcal{N} \left(\pmb{G}_0 \left\{\Bar{\pmb{s}}(t_0)+\pmb{\varphi}(t_0)\right\}, \pmb{G}_0\pmb{\Psi}(t_0)\pmb{G}_0^\top+\pmb{\Sigma}_0 \right). \label{dist_z0}
\end{equation}
Combining the LNA prior of $\pmb{s}(t_0)$ with \eqref{dist_z0}, we obtain the joint distribution of $\pmb{s}(t_0)$ and $\pmb{y}_0(t_0)$ as
\begin{equation*}
    \begin{pmatrix}
        \pmb{s}(t_0) \\
        \pmb{y}_0(t_0)
    \end{pmatrix} \bigg| \pmb{\sigma} \sim \mathcal{N} \left\{
    \begin{pmatrix}
        \Bar{\pmb{s}}(t_0)+\pmb{\varphi}(t_0) \\
        \pmb{G}_0 \left\{\Bar{\pmb{s}}(t_0)+\pmb{\varphi}(t_0)\right\}
    \end{pmatrix},
    \begin{pmatrix}
        \pmb{\Psi}(t_0) & \pmb{\Psi}(t_0)\pmb{G}_0^\top \\
        \pmb{G}_0\pmb{\Psi}(t_0) & \pmb{G}_0\pmb{\Psi}(t_0)\pmb{G}_0^\top+\pmb{\Sigma}_0
    \end{pmatrix}
    \right\}.
\end{equation*}
By applying the conditional distribution properties of multivariate Gaussian distribution, the posterior distribution of $\pmb{s}(t_0)$ is updated based on the observation $\pmb{y}_0(t_0)$, i.e.,
\begin{equation}
    \pmb{s}(t_0) | \pmb{y}_0(t_0);\pmb{\sigma} \sim \mathcal{N} \left(\pmb{\alpha}(t_0), \pmb{\beta}(t_0)\right), \label{posterior_s_t0}
\end{equation}
where
\begin{align}
    \pmb{\alpha}(t_0) &= \Bar{\pmb{s}}(t_0)+\pmb{\varphi}(t_0) + \pmb{\Psi}(t_0)\pmb{G}_0^\top \left(\pmb{G}_0\pmb{\Psi}(t_0)\pmb{G}_0^\top+\pmb{\Sigma}_0\right)^{-1} \left(\pmb{y}_0(t_0)-\pmb{G}_0 \left\{\Bar{\pmb{s}}(t_0)+\pmb{\varphi}(t_0)\right\}\right), \label{alpha_t0} \\
    \pmb{\beta}(t_0) &= \pmb{\Psi}(t_0) -  \pmb{\Psi}(t_0)\pmb{G}_0^\top \left(\pmb{G}_0\pmb{\Psi}(t_0)\pmb{G}_0^\top+\pmb{\Sigma}_0\right)^{-1}  \pmb{G}_0\pmb{\Psi}(t_0). \label{beta_t0}
\end{align}

\textbf{Step 2:} For the subsequent observation time points $t_1,t_2,\ldots,t_H$, we apply the idea of Kalman filter to sequentially update the LNA prior for $\pmb{s}(t_{h+1})$ and calculate the approximate $p\left(\pmb{y}_{h+1}(t_{h+1}) | \pmb{y}_h(t_h),\ldots,\pmb{y}_0(t_0);\pmb{\theta},\pmb{\sigma}\right)$ recursively for $h=0,1,\ldots,H-1$. Specifically, we first reinitialize the initial values of the ODEs \eqref{ODE} and \eqref{SDE_cov} to the posterior mean and covariance of $\pmb{s}(t_h)$ respectively. That is, set $\Bar{\pmb{s}}(t_h) = \pmb{\alpha}(t_h)$ and $\pmb{\Psi}(t_h) = \pmb{\beta}(t_h)$. We let $\pmb{\varphi}(t_h) = 0$ as $\pmb{\varphi}(t_k) = 0$ for all $k \geq h$ according to the ODE \eqref{SDE_mean}. By integrating the ODEs \eqref{ODE} and \eqref{SDE_cov} through time $t_h$ to $t_{h+1}$, we obtain $\Bar{\pmb{s}}(t_{h+1})$ and $\pmb{\Psi}(t_{h+1})$. In practice we work with their discretized versions, given by the Euler method, 
\begin{align}
    \Bar{\pmb{s}}(t_{h+1}) &= \Bar{\pmb{s}}(t_h) + \pmb{\mu}(\Bar{\pmb{s}}(t_h);\pmb{\theta}) \Delta t_h, 
    \label{ODE_discrete}\\
    \pmb{\Psi}(t_{h+1}) &= \pmb{\Psi}(t_h) + \left\{ \pmb{\Psi}(t_h) \left(\nabla_{\pmb{s}} \pmb{\mu}(\pmb{s};\pmb{\theta})|_{\pmb{s}=\Bar{\pmb{s}}(t_h)}\right)^\top + \nabla_{\pmb{s}} \pmb{\mu}(\pmb{s};\pmb{\theta})|_{\pmb{s}=\Bar{\pmb{s}}(t_h)} \pmb{\Psi}(t_h) + \pmb{D}(\Bar{\pmb{s}}(t_h);\pmb{\theta}) \right\} \Delta t_h. 
    \label{SDE_cov_discrete}
\end{align}
As $\Delta t_h = t_{h+1}-t_h$ is often too large to be used as a time step in \eqref{ODE_discrete} and \eqref{SDE_cov_discrete}, we introduce $\Delta z_h = \Delta t_h/I_h$ for some positive integer $I_h \geq 1$. By choosing $I_h$ to be sufficiently large, we can ensure the discretization error associated with the Euler method is arbitrarily small. That is, to compute $\Bar{\pmb{s}}(t_{h+1})$ and $\pmb{\Psi}(t_{h+1})$ more accurately, we recursively calculate the following equations for $i=0,1,\ldots,I_h-1$,
\begin{align}
    \Bar{\pmb{s}}(t_h+(i+1)\Delta z_h) &= \Bar{\pmb{s}}(t_h+i\Delta z_h) + \pmb{\mu}(\Bar{\pmb{s}}(t_h+i\Delta z_h);\pmb{\theta}) \Delta z_h, \label{ODE_discrete_recur} \\
    \pmb{\Psi}(t_h+(i+1)\Delta z_h) &= \pmb{\Psi}(t_h+i\Delta z_h) + \left\{ \pmb{\Psi}(t_h+i\Delta z_h) \left(\nabla_{\pmb{s}} \pmb{\mu}(\pmb{s};\pmb{\theta})|_{\pmb{s}=\Bar{\pmb{s}}(t_h+i\Delta z_h)}\right)^\top + \right. \nonumber \\
    &\quad \left. \nabla_{\pmb{s}} \pmb{\mu}(\pmb{s};\pmb{\theta})|_{\pmb{s}=\Bar{\pmb{s}}(t_h+i\Delta z_h)} \pmb{\Psi}(t_h+i\Delta z_h) + \pmb{D}(\Bar{\pmb{s}}(t_h+i\Delta z_h);\pmb{\theta}) \right\} \Delta z_h. \label{SDE_cov_discrete_recur}
\end{align}
Therefore, we get the updated LNA prior on $\pmb{s}(t_{h+1})$ by applying \eqref{dist_LNA}, i.e.,
\begin{equation}
    \pmb{s}(t_{h+1}) | \pmb{y}_h(t_h),\ldots,\pmb{y}_0(t_0);\pmb{\theta},\pmb{\sigma} \sim \mathcal{N} \left(\Bar{\pmb{s}}(t_{h+1}), \pmb{\Psi}(t_{h+1})\right). \label{dist_s}
\end{equation}
Here, LNA gives us a Gaussian approximation to the transition density from $\pmb{s}(t_h) | \pmb{y}_h(t_h),\ldots,\pmb{y}_0(t_0);\pmb{\theta},\pmb{\sigma}$ to $\pmb{s}(t_{h+1}) | \pmb{y}_h(t_h),\ldots,\pmb{y}_0(t_0);\pmb{\theta},\pmb{\sigma}$. Then, based on the model of measurement uncertainty or error in \eqref{measurement error}, we get a one-step forecast of the observation $\pmb{y}_{h+1}(t_{h+1})$ as
\begin{equation}
    \pmb{y}_{h+1}(t_{h+1}) | \pmb{y}_h(t_h),\ldots,\pmb{y}_0(t_0);\pmb{\theta},\pmb{\sigma} \sim \mathcal{N} \left(\pmb{G}_{h+1}\Bar{\pmb{s}}(t_{h+1}), \pmb{G}_{h+1}\pmb{\Psi}(t_{h+1})\pmb{G}_{h+1}^\top+\pmb{\Sigma}_{h+1}\right). \label{dist_y_h+1}
\end{equation}
Combining the distributions \eqref{dist_s} and \eqref{dist_y_h+1}, we obtain the joint distribution as
\begin{equation*}
    \begin{pmatrix}
        \pmb{s}(t_{h+1}) \\
        \pmb{y}_{h+1}(t_{h+1})
    \end{pmatrix} \bigg| \pmb{y}_h(t_h),\ldots,\pmb{y}_0(t_0);\pmb{\theta},\pmb{\sigma} \sim \mathcal{N} \left\{
    \begin{pmatrix}
        \Bar{\pmb{s}}(t_{h+1}) \\
        \pmb{G}_{h+1}\Bar{\pmb{s}}(t_{h+1})
    \end{pmatrix},
    \begin{pmatrix}
        \pmb{\Psi}(t_{h+1}) & \pmb{\Psi}(t_{h+1})\pmb{G}_{h+1}^\top \\
        \pmb{G}_{h+1}\pmb{\Psi}(t_{h+1}) & \pmb{G}_{h+1}\pmb{\Psi}(t_{h+1})\pmb{G}_{h+1}^\top+\pmb{\Sigma}_{h+1}
    \end{pmatrix}
    \right\}.
\end{equation*}
Thus, the posterior distribution of $\pmb{s}(t_{h+1})$ becomes
\begin{equation}
    \pmb{s}(t_{h+1}) | \pmb{y}_{h+1}(t_{h+1}),\ldots,\pmb{y}_0(t_0);\pmb{\theta},\pmb{\sigma} \sim \mathcal{N} \left(\pmb{\alpha}(t_{h+1}), \pmb{\beta}(t_{h+1})\right), \label{posterior_s_th}
\end{equation}
where
\begin{align}
    \pmb{\alpha}(t_{h+1}) &= \Bar{\pmb{s}}(t_{h+1}) + \pmb{\Psi}(t_{h+1})\pmb{G}_{h+1}^\top \left(\pmb{G}_{h+1}\pmb{\Psi}(t_{h+1})\pmb{G}_{h+1}^\top+\pmb{\Sigma}_{h+1}\right)^{-1} \left(\pmb{y}_{h+1}(t_{h+1})-\pmb{G}_{h+1}\Bar{\pmb{s}}(t_{h+1})\right), \label{alpha_th}\\
    \pmb{\beta}(t_{h+1}) &= \pmb{\Psi}(t_{h+1}) - \pmb{\Psi}(t_{h+1})\pmb{G}_{h+1}^\top \left(\pmb{G}_{h+1}\pmb{\Psi}(t_{h+1})\pmb{G}_{h+1}^\top+\pmb{\Sigma}_{h+1}\right)^{-1} \pmb{G}_{h+1}\pmb{\Psi}(t_{h+1}). \label{beta_th}
\end{align}

\textbf{Step 3:} From the distributions \eqref{dist_z0} and \eqref{dist_y_h+1} for $h=0,1,\ldots,H-1$, the likelihood of the observations $\mathcal{D}_M$ can be calculated by the following decomposition,
\begin{equation}
    p\left(\pmb{y}_0(t_0),\pmb{y}_1(t_1),\ldots,\pmb{y}_H(t_H) | \pmb{\theta},\pmb{\sigma}\right) = p\left(\pmb{y}_0(t_0) | \pmb{\theta},\pmb{\sigma}\right) \prod_{h=0}^{H-1} p\left(\pmb{y}_{h+1}(t_{h+1}) | \pmb{y}_h(t_h),\ldots,\pmb{y}_0(t_0);\pmb{\theta},\pmb{\sigma}\right). \label{eq:likelihood}
\end{equation}

\section{BAYESIAN ANALYSIS AND Algorithm Development}
\label{sec:BayesianInference}

In this section, we simultaneously infer the model parameters $\pmb{\theta}$ and the measurement error level $\pmb{\sigma}$ from the observations $\mathcal{D}_M$. By applying the Bayes' rule, we have the joint posterior distribution of $\pmb{\theta}$ and $\pmb{\sigma}$,
\begin{align}
    p\left(\pmb{\theta},\pmb{\sigma} | \mathcal{D}_M\right) &\propto  
    p\left(\pmb{\theta}\right) p\left(\pmb{\sigma}\right) \exp \left\{ -\frac{1}{2}\left[M|\pmb{J}_0| \log(2\pi) + \log \left|\pmb{G}_0\pmb{\Psi}(t_0)\pmb{G}_0^\top+\pmb{\Sigma}_0\right|  \right. \right. \nonumber \\ 
    &  \quad \left. +\left(\pmb{y}_0(t_0)-\pmb{G}_0 \left\{\Bar{\pmb{s}}(t_0)+\pmb{\varphi}(t_0)\right\}\right)^\top \left(\pmb{G}_0\pmb{\Psi}(t_0)\pmb{G}_0^\top+\pmb{\Sigma}_0\right)^{-1} \left(\pmb{y}_0(t_0)-\pmb{G}_0 \left\{\Bar{\pmb{s}}(t_0)+\pmb{\varphi}(t_0)\right\}\right)\right]  \nonumber \\
    &\quad -\frac{1}{2} \sum_{h=0}^{H-1} \left[M|\pmb{J}_{h+1}| \log(2\pi) + \log \left|\pmb{G}_{h+1}\pmb{\Psi}(t_{h+1})\pmb{G}_{h+1}^\top+\pmb{\Sigma}_{h+1}\right|  \right. \nonumber \\ 
    &\quad \left. \left. + \left(\pmb{y}_{h+1}(t_{h+1})-\pmb{G}_{h+1}\Bar{\pmb{s}}(t_{h+1})\right)^\top\left(\pmb{G}_{h+1}\pmb{\Psi}(t_{h+1})\pmb{G}_{h+1}^\top+\pmb{\Sigma}_{h+1}\right)^{-1} \left(\pmb{y}_{h+1}(t_{h+1})-\pmb{G}_{h+1}\Bar{\pmb{s}}(t_{h+1})\right)\right] \right\}, \label{eq:joint_posterior}
\end{align}
where $p\left(\pmb{\theta}\right)$ and $ p\left(\pmb{\sigma}\right)$ are the priors for $\pmb{\theta}$ and $\pmb{\sigma}$ respectively. By utilizing the joint posterior distribution $p\left(\pmb{\theta},\pmb{\sigma} | \mathcal{D}_M\right)$ in \eqref{eq:joint_posterior}, we further develop a MALA procedure to efficiently generate posterior samples of the $L$-dimensional parameters $\pmb{\eta} \equiv\left(\pmb{\theta}^\top,\pmb{\sigma}^\top\right)^\top$ where $L=N+|\pmb{J}_y|$.

By utilizing the gradients of posterior (\ref{eq:joint_posterior}), MALA generates more promising candidate samples at the parameter space with higher posterior probability. It improves the mixing of classic MCMC algorithm through utilizing a combination of two mechanisms, i.e., Langevin diffusion and Metropolis-Hastings step. Langevin diffusion is originally a gradient descent of a potential function (representing a force field in physics) plus a Brownian motion term accounting for thermodynamics. To overcome the limitation of random walk-based search strategies used in classic MCMC, we leverage on the information provided by the closed form posterior distribution $p\left( \pmb{\eta} | \mathcal{D}_M\right)$ and use Langevin diffusion to develop a more efficient posterior sampling approach. We construct a continuous-time stochastic process characterizing the Langevin diffusion-based posterior search. Specifically, we consider the following (overdamped) Langevin diffusion
\begin{equation}
    d\pmb{\eta}(\tau) = \nabla_{\pmb{\eta}} \log p\left( \pmb{\eta} | \mathcal{D}_M\right) |_{\pmb{\eta} = \pmb{\eta}(\tau)} d\tau + \sqrt{2} d\pmb{W}(\tau) \label{Langevin diffusion}
\end{equation}
driven by the time derivative of an $L$-dimensional standard Brownian motion (i.e., $d\pmb{W}(\tau)$). It speeds up the MCMC convergence through drifting the search with the gradient of the target log-posterior distribution (i.e., $\log p( \pmb{\eta} | \mathcal{D}_M)$), which drives the random walk towards the parameter region with high posterior probability.

To numerically solve Equation \eqref{Langevin diffusion} and generate posterior samples from $p( \pmb{\eta} | \mathcal{D}_M)$, the Euler-Maruyama approximation \cite{kloeden1992numerical} is used to obtain the discretized Langevin diffusion with a step size $\Delta \tau > 0$, 
\begin{equation}
    \pmb{\eta}(\tau+1) := \pmb{\eta}(\tau) + \nabla_{\pmb{\eta}} \log p\left( \pmb{\eta} | \mathcal{D}_M\right) |_{\pmb{\eta} = \pmb{\eta}(\tau)} \Delta \tau + \sqrt{2} \Delta\pmb{W}(\tau), \label{update rule}
\end{equation}
where each $\Delta \pmb{W}(\tau) \in \mathbb{R}^L$ is a Gaussian random vector with mean zero and covariance diag$\{\Delta \tau\} \in \mathbb{R}^{L \times L}$. The gradient of the log-posterior 
\begin{align*}
    \nabla_{\pmb{\eta}} \log p\left( \pmb{\eta} | \mathcal{D}_M\right) 
    &=\left(\left\{\frac{\partial \log p\left(\pmb{\theta},\pmb{\sigma} | \mathcal{D}_M\right)}{\partial \theta_n}, n \in [N]\right\}, \left\{\frac{\partial \log p\left(\pmb{\theta},\pmb{\sigma} | \mathcal{D}_M\right)}{\partial \sigma_{jj}}, j \in \pmb{J}_y\right\}\right)^\top
\end{align*}
is tractable from Equation \eqref{eq:joint_posterior}. In particular, we provide a recursive procedure in Algorithm \ref{Algr:posterior} to compute $p\left(\pmb{\eta} | \mathcal{D}_M\right)$ and $\nabla_{\pmb{\eta}} \log p\left( \pmb{\eta} | \mathcal{D}_M\right)$. 

\begin{algorithm}[th]
\DontPrintSemicolon
\KwIn{The priors $p\left(\pmb{\theta}\right)$ and $p\left(\pmb{\sigma}\right)$, observations $\mathcal{D}_M = \{\pmb{y}_h(t_h)\}_{h=0}^{H}$, ODE initial values $\Bar{\pmb{s}}(t_0)$, $\pmb{\varphi}(t_0)$ and $\pmb{\Psi}(t_0)$, constant matrices $\pmb{G}_h$, and appropriate positive integers $I_h$ for $h=0,1,\ldots,H-1$.}
\KwOut{$p\left(\pmb{\eta} | \mathcal{D}_M\right)$ and $\nabla_{\pmb{\eta}} \log p\left( \pmb{\eta} | \mathcal{D}_M\right)$.}
\textbf{1.} Calculate $\pmb{\alpha}(t_0)$ and $\pmb{\beta}(t_0)$ by applying Equations \eqref{alpha_t0} and \eqref{beta_t0};\\
\textbf{2.} Calculate $\partial \pmb{\alpha}(t_0)/\partial \theta_n$, $\partial \pmb{\beta}(t_0)/\partial \theta_n$ for $n \in [N]$, and $\partial \pmb{\alpha}(t_0)/\partial \sigma_{jj}$, $\partial \pmb{\beta}(t_0)/\partial \sigma_{jj}$ for $j \in \pmb{J}_y$;\\
\For{$h=0,1,\ldots,H-1$}{
    \For{$i=0,1,\ldots,I_h-1$}{\textbf{3.} Calculate $\Bar{\pmb{s}}(t_h+(i+1)\Delta z_h)$ and $\pmb{\Psi}(t_h+(i+1)\Delta z_h)$ by applying Equations \eqref{ODE_discrete_recur} and \eqref{SDE_cov_discrete_recur};\\
    \textbf{4.} Calculate $\partial \Bar{\pmb{s}}(t_h+(i+1)\Delta z_h)/\partial \theta_n$, $\partial \pmb{\Psi}(t_h+(i+1)\Delta z_h)/\partial \theta_n$ for $n \in [N]$, and $\partial \Bar{\pmb{s}}(t_h+(i+1)\Delta z_h)/\partial \sigma_{jj}$, $\partial \pmb{\Psi}(t_h+(i+1)\Delta z_h)/\partial \sigma_{jj}$ for $j \in \pmb{J}_y$;
    }
    \textbf{5.} Calculate $\pmb{\alpha}(t_{h+1})$ and $\pmb{\beta}(t_{h+1})$ by applying Equations \eqref{alpha_th} and \eqref{beta_th};\\
    \textbf{6.} Calculate $\partial \pmb{\alpha}(t_{h+1})/\partial \theta_n$, $\partial \pmb{\beta}(t_{h+1})/\partial \theta_n$ for $n \in [N]$, and $\partial \pmb{\alpha}(t_{h+1})/\partial \sigma_{jj}$, $\partial \pmb{\beta}(t_{h+1})/\partial \sigma_{jj}$ for $j \in \pmb{J}_y$;\\
}
    \textbf{7.} Return $p\left(\pmb{\eta} | \mathcal{D}_M\right)$ by applying Equation \eqref{eq:joint_posterior}, and $\nabla_{\pmb{\eta}} \log p\left( \pmb{\eta} | \mathcal{D}_M\right)$ by calculating $\partial \log p\left(\pmb{\eta} | \mathcal{D}_M\right)/\partial \theta_n$ for $n \in [N]$ and $\partial \log p\left(\pmb{\eta} | \mathcal{D}_M\right)/\partial \sigma_{jj}$ for $j \in \pmb{J}_y$.
\caption{Computing $p\left(\pmb{\eta} | \mathcal{D}_M\right)$ and $\nabla_{\pmb{\eta}} \log p\left( \pmb{\eta} | \mathcal{D}_M\right)$.}
\label{Algr:posterior}   
\end{algorithm}

To correct the bias in the stationary distribution induced by the discretization used in the update rule \eqref{update rule}, a Metropolis-Hastings step is incorporated for simulating the Langevin diffusion \eqref{Langevin diffusion}. Specifically, we consider the update rule \eqref{update rule} and define a proposal distribution to generate a new MCMC posterior sample $\Tilde{\pmb{\eta}}(\tau+1)$,
\begin{equation}
    \Tilde{\pmb{\eta}}(\tau+1) := \pmb{\eta}(\tau) + \nabla_{\pmb{\eta}} \log p\left( \pmb{\eta} | \mathcal{D}_M\right) |_{\pmb{\eta} = \pmb{\eta}(\tau)} \Delta \tau + \sqrt{2} \Delta\pmb{W}(\tau). \label{proposal}
\end{equation}
Thus, the MCMC conditional sampling distribution $\Tilde{\pmb{\eta}}(\tau+1) | \pmb{\eta}(\tau)$ is Gaussian distributed with mean $\pmb{\eta}(\tau) + \nabla_{\pmb{\eta}} \log p\left( \pmb{\eta} | \mathcal{D}_M\right) |_{\pmb{\eta} = \pmb{\eta}(\tau)} \Delta \tau$ and covariance diag$\{2\Delta \tau\} \in \mathbb{R}^{L \times L}$. Then, the candidate sample $\Tilde{\pmb{\eta}}(\tau+1)$ generated from this proposal is accepted with the ratio,
\begin{equation}
    \gamma_{\rm acc} := \min \left\{1, \frac{p\left(\Tilde{\pmb{\eta}}(\tau+1) | \mathcal{D}_M\right)q_{\rm trans}\left(\pmb{\eta}(\tau)\bigg|\Tilde{\pmb{\eta}}(\tau+1)\right)}{p\left(\pmb{\eta}(\tau) | \mathcal{D}_M\right)q_{\rm trans}\left(\Tilde{\pmb{\eta}}(\tau+1)\bigg|\pmb{\eta}(\tau)\right)}\right\}, \label{acceptance ratio}
\end{equation}
where the proposal distribution $q_{\rm trans}(\pmb{\eta}^\prime|\pmb{\eta}) \propto \exp \left\{-\frac{1}{4\Delta \tau} \bigg|\bigg|\pmb{\eta}^\prime - \pmb{\eta} - \nabla_{\pmb{\eta}} \log p\left(\pmb{\eta} | \mathcal{D}_M\right)\Delta \tau\bigg|\bigg|^2\right\}$ ($||\cdot||$ denotes the Euclidean norm) is the transition density from $\pmb{\eta}$ to $\pmb{\eta}^\prime$ obtained from Equation \eqref{proposal}.

In sum, we provide the MALA procedure in Algorithm \ref{Algr:MALA} to generate posterior samples for the enzymatic SRN mechanistic model parameters $\pmb{\theta}$ and the measurement error level $\pmb{\sigma}$ together. Within each $\tau$-th iteration of MALA joint posterior sampler, given the previous sample $\pmb{\eta}(\tau)$, we compute and generate one proposal sample $\Tilde{\pmb{\eta}}(\tau+1)$ from the discretized Langevin diffusion \eqref{proposal}, and accept it with the Metropolis-Hastings ratio \eqref{acceptance ratio}. By repeating this procedure, $\pmb{\theta}$ and $\pmb{\sigma}$ are updated together with a joint gradient at each iteration, and we thus get samples $\pmb{\eta}(\tau) = \left(\pmb{\theta}^\top(\tau),\pmb{\sigma}^\top(\tau)\right)^\top$ with $\tau = 1,2,\ldots,T_0+(B-1)\delta$. To reduce the initial bias and correlations between consecutive samples, we discard an appropriate burn-in period for convergence (i.e., the first $T_0$ samples) and then keep one for every $\delta$ samples. Consequently, we obtain the posterior samples $\pmb{\eta}^{(b)} \sim p\left(\pmb{\eta}|\mathcal{D}_M\right)$ with $b=1,2,\ldots,B$. 

\begin{algorithm}[th]
\DontPrintSemicolon
\KwIn{The priors $p\left(\pmb{\theta}\right)$ and $p\left(\pmb{\sigma}\right)$, step size $\Delta \tau$, posterior sample size $B$, initial warm-up length $T_0$, and an appropriate integer $\delta$ to reduce sample correlation.}
\KwOut{Posterior samples $\pmb{\eta}^{(b)} \sim p(\pmb{\eta}|\mathcal{D}_M)$ with $b=1,2,\ldots,B$.}
\textbf{1.} Set the initial values $\pmb{\eta}(0) := \left(\pmb{\theta}^\top(0),\pmb{\sigma}^\top(0)\right)^\top$ by sampling from the priors;\\
\For{$\tau=0,1,\ldots,T_0+(B-1)\delta$}{
    \textbf{2.} Calculate $p\left(\pmb{\eta}(\tau) | \mathcal{D}_M\right)$ and $\nabla_{\pmb{\eta}} \log p\left( \pmb{\eta} | \mathcal{D}_M\right) |_{\pmb{\eta} = \pmb{\eta}(\tau)}$ by calling Algorithm \ref{Algr:posterior};\\
    \textbf{3.} Generate a proposal $\Tilde{\pmb{\eta}}(\tau+1)$ by applying Equation \eqref{proposal};\\
    \textbf{4.} Calculate the acceptance ratio $\gamma_{\rm acc}$ by applying Equation \eqref{acceptance ratio};\\
    \textbf{5.} Draw $u$ from the continuous uniform distribution $U(0,1)$;\\
    \If{$u \leq \gamma_{\rm acc}$}{
    \textbf{6.} The proposal $\Tilde{\pmb{\eta}}(\tau+1)$ is accepted, and set $\pmb{\eta}(\tau+1):=\Tilde{\pmb{\eta}}(\tau+1)$;}
    \ElseIf{$u > \gamma_{\rm acc}$}{
    \textbf{7.} The proposal $\Tilde{\pmb{\eta}}(\tau+1)$ is rejected, and set $\pmb{\eta}(\tau+1):=\pmb{\eta}(\tau)$;}}
\textbf{8.} Return posterior samples $\pmb{\eta}^{(b)} := \pmb{\eta}(T_0+(b-1)\delta+1)$ for $b=1,2,\ldots,B$. 
\caption{MALA joint posterior sampler for SRN.}
\label{Algr:MALA}   
\end{algorithm}

\begin{remark}
    From Equation \eqref{proposal}, the support of the posterior samples $\pmb{\eta}^{(b)}$ generated by Algorithm \ref{Algr:MALA} is the entire $L$-dimensional real space $\mathbb{R}^{L}$. But in most real-word cases including SRN, the feasible space of $\pmb{\eta}$ is restricted, meaning it can be a subset of $\mathbb{R}^{L}$. For instance, some biological parameters such as rates should be ensured positivity, while some parameters such as probabilities or bioavailability should be between 0 and 1 \cite{prague2013nimrod}. Reparametrization of the system allows us to take these constraints into account. Specifically, we can introduce one-to-one functions $f_l(\cdot)$ for $l=1,2,\ldots,L$, and define transformed parameters $\eta_{l}^{\rm trans} = f_l(\eta_{l})$. For instance, $f_l(\cdot)$ can be logarithmic functions to transform the support from the positive space to the real space, or inverse Logistic functions to transform the support from the interval $[0,1]$ to the real space. Then we can perform Algorithm \ref{Algr:MALA} on the transformed $\pmb{\eta}^{\rm trans} = \left(\eta_{1}^{\rm trans},\eta_{2}^{\rm trans},\ldots,\eta_{L}^{\rm trans}\right)^\top$. 
\end{remark}

\section{EMPIRICAL STUDY}
\label{sec:empirical}

In this section, we use a representative example of SRN, i.e., Michaelis-Menten enzyme kinetics \cite{rao2003stochastic}, to assess the empirical performance of the proposed Bayesian inference approach. In specific, we consider the Michaelis-Menten enzyme kinetic model involving four biochemical species, i.e., Enzyme, Substrate, Complex, and Product. It describes the catalytic conversion of a substrate into a product via an enzymatic reaction involving enzyme, represented by the following three chemical reactions,
\begin{flalign*}
    && & \begin{array}{l}
        \text{Reaction } 1: \ {\rm Enzyme} + {\rm Substrate} \longrightarrow {\rm Complex}, \\
        \text{Reaction } 2: \ {\rm Complex} \longrightarrow {\rm Enzyme} + {\rm Substrate}, \\
        \text{Reaction } 3: \ {\rm Complex} \longrightarrow {\rm Enzyme} + {\rm Product},
    \end{array}
    & ~~~ \mbox{with} ~~~ \pmb{C} = 
    \begin{pmatrix}
        -1 & 1 & 1 \\
        -1 & 1 & 0 \\
        1 & -1 & -1 \\
        0 & 0 & 1 
    \end{pmatrix}. &&&&
\end{flalign*}
In particular, let $\pmb{s}(t) = \left(s_1(t),s_2(t),s_3(t),s_4(t)\right)^\top$ denote the system state vector at any time $t$, where $s_1(t)$, $s_2(t)$, $s_3(t)$, and $s_4(t)$ are the respective concentration of Enzyme, Substrate, Complex, and Product. The stoichiometry matrix $\pmb{C}$ associated with the system can be obtained from the above three reaction equations, and the associated reaction rate vector is $\pmb{v}(\pmb{s}(t);\pmb{\theta}) = \left(\theta_1 s_1(t) s_2(t), \theta_2 s_3(t), \theta_3 s_3(t)\right)^\top$. Our goal is to perform Bayesian inference for the vector of the unknown kinetic rate parameters $\pmb{\theta} = \left(\theta_1,\theta_2,\theta_3\right)^\top$.

We simulate a synthetic dataset for 80 seconds (i.e., on the time interval $[0,80]$ seconds) using the Gillespie algorithm \cite{gillespie1977exact} to ensure exact simulation with the true parameters $\pmb{\theta}^{\rm true} = \left(0.001,0.005,0.01\right)^\top$, and the initial states $\pmb{s}(0) = \left(45,39,55,6\right)^\top$. These initial values are  obtained after running the process for a short time from some arbitrarily chosen population levels. And we create a challenging data-poor scenario for model inference by assuming incomplete and noisy observations. Specifically, we consider one batch size (i.e., $M=1$), and discard observations on the Enzyme, Substrate, and Product levels, and only the Complex level is observed every $\Delta t$ seconds from $t_0 = 0$ to $t_{H} = 80$ ($H + 1$ observation time points in total), so that $\pmb{J}_h = \{3\}$ and $\pmb{G}_h = \left(0,0,1,0\right)$ for any $h=0,1,\ldots,H$. And we assume that there is homogeneous additive Gaussian measurement error, i.e., $\epsilon(t_h) \sim \mathcal{N}(0, \sigma)$ where $\sigma = 4$; that is, the error standard deviation is two Complex molecules. The inferred parameter vector is $\pmb{\eta} \equiv (\pmb{\theta}^\top, \sigma)^\top$.

We assess the performance of the proposed joint posterior sampler under model uncertainty induced with the different data sizes, i.e., $H = 4, 8, 16$ ($\Delta t = 20, 10, 5$ seconds correspondingly). To study the effect of additional gradient information and Bayesian updating step, the MALA with Bayesian updating LNA is compared to other candidate approaches, including (1) classic Metropolis-Hastings algorithm (M-H) with Bayesian updating LNA, and (2) MALA with original LNA (without Bayesian update), in terms of convergence behavior of posterior sampling. Since the support of the parameters is the positive space, we first need to use the logarithmic function to transform it to the real space. That is, we set $\log(\pmb{\eta}) = \left(\log(\theta_1),\log(\theta_2),\log(\theta_3),\log(\sigma)\right)^\top$ and apply three algorithms to $\log(\pmb{\eta})$. For both LNA metamodels, we set $\Bar{\pmb{s}}(t_0)+\pmb{\varphi}(t_0)=\left(50,40,60,10\right)^\top$, $\pmb{\Psi}(t_0)$ as a 4-by-4 identity matrix, and $I_h = 2000, 1000, 500$ for $\Delta t = 20, 10, 5$ respectively to make a small $\Delta z_h = 0.01$ for any $h=0,1,\ldots,H$. The priors of the parameters are set as $\theta_k \sim U(0,1)$ for $k=1,2,3$, and $\sigma \sim U(0,25)$, to consider a difficult situation without strong prior information. The results are estimated based on 10 macro-replications.

\begin{figure}[htbp]
    \centering
    \subfloat[$\log(\theta_1)$]{\includegraphics[width=0.34\columnwidth]{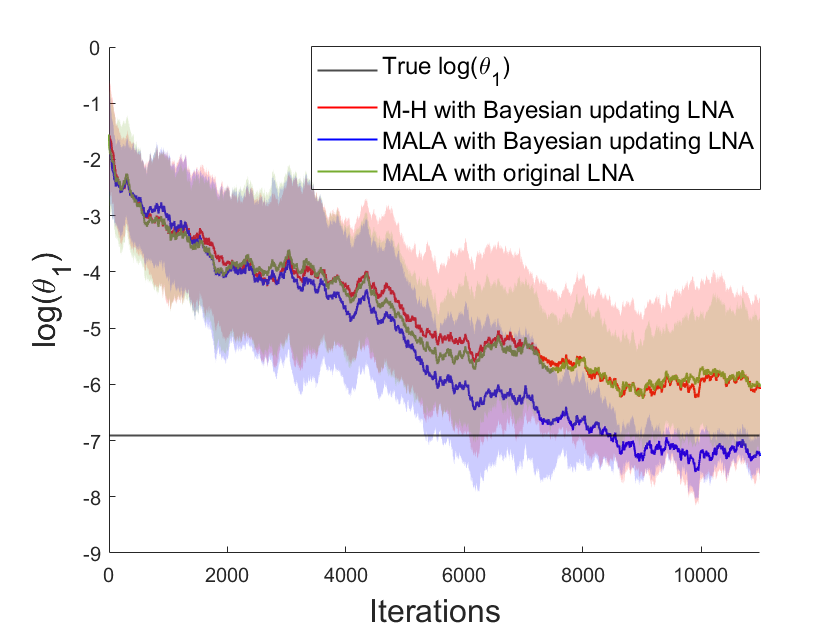}} 
    \subfloat[$\log(\theta_2)$]{\includegraphics[width=0.34\columnwidth]{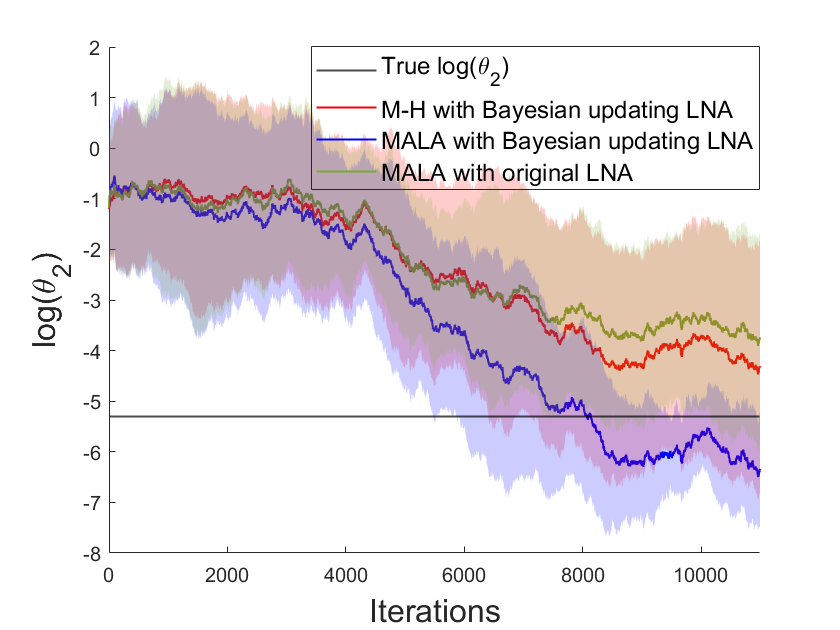}} 
    \subfloat[$\log(\theta_3)$]{\includegraphics[width=0.34\columnwidth]{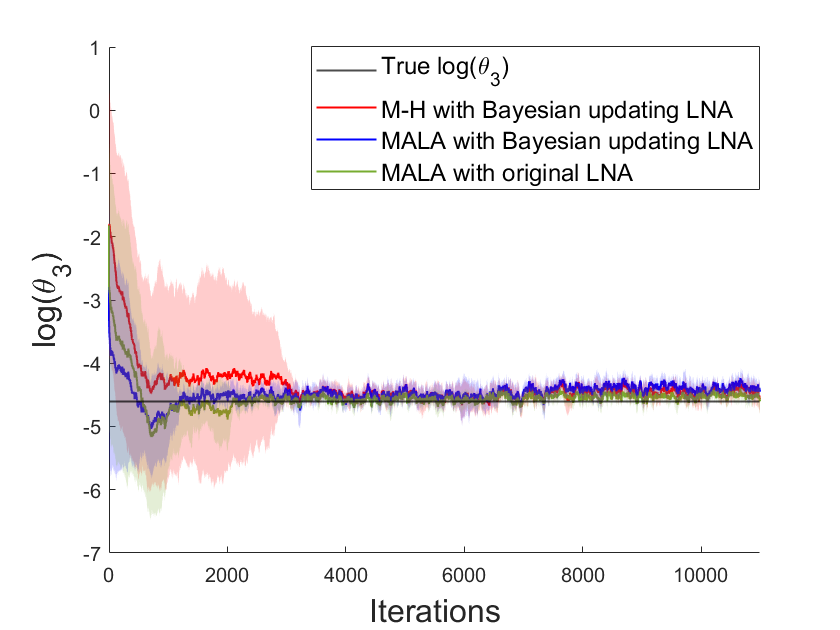}} 
    \caption{{The convergence trends of (1) MALA with original LNA, (2) MALA with Bayesian updating LNA, and (3) M-H with Bayesian updating LNA (with 95\% CIs) when the data size $H=16 \ (\Delta t=5)$.}}
    \label{fig:parameter convergence}
\end{figure}

First, we compare the convergence speed of three algorithms. For MALA with Bayesian updating LNA and with original LNA (without Bayesian update), we set the step size $\Delta \tau = 0.001$. Correspondingly, to show that MALA improves the mixing of MCMC, for M-H with Bayesian updating LNA, we set its proposal distribution to be Gaussian distributed with the current sample as mean and diag$\{2\Delta \tau\}=$ diag$\{0.002\}$ as covariance. Figure \ref{fig:parameter convergence} shows the mean convergence trends of the three algorithms for the three log-kinetic rate parameters (with 95\% confidence intervals (CIs) across 10 macro-replications) under the data size $H=16 \ (\Delta t=5)$. The black line represents the true log-parameters. By comparing the widths of the CIs as iterations progress, we observe that MALA shows a significant improvement in the convergence speed of the log-kinetic rate parameters inference over M-H, while the Bayesian updating step reduces the approximation error accumulation over time of original LNA, providing a more accurate likelihood approximation. It demonstrates that by sufficiently leveraging on the likelihood and its gradient information provided by a moderate size of observations, MALA posterior sampling based on the likelihood approximated by the Bayesian updating LNA metamodel converges quickly towards the true log-parameters.

Then, we study the root mean square error (RMSE) of model parameter estimation to assess the convergence results of three algorithms under three different data sizes. Basically, the RMSE measures the differences between the true and the estimated log-parameters based on $B$ posterior samples, i.e., RMSE $= \sqrt{\frac{1}{B} \sum_{b=1}^B |\log(\eta_l^{\rm true})-\{\log(\eta_l)\}^{(b)}|^2}$ for $l=1,2,3,4$. We set the initial warm-up length $T_0=10000$ to reduce the initial bias, the posterior sample size $B=100$, and an appropriate integer $\delta=10$ to reduce sample correlation for three algorithms. We summarize the 95\% CIs obtained by using 10 macro-replications of the RMSEs for the four log-parameters inferred by the three algorithms in Table \ref{table:rmse}. As it shows, for the four log-parameters except $\log (\theta_3)$ inferred under the data size $H=8,16$, the RMSEs of MALA with Bayesian updating LNA decrease more significantly than those of MALA with original LNA as the data size increases, demonstrating the major benefit induced by the Bayesian updating LNA metamodel. That means it refines the approximation of the likelihood as more observations available to update the original LNA model. The exception of $\log (\theta_3)$ is due to that, $\theta_3$ is the kinetic rate parameter associated with the Reaction 3, whose reactant (i.e., Complex) is the only observable component of this system, the data size of $H=16$ is thus sufficient to provide relatively accurate likelihood information based on the original LNA metamodel, while the Bayesian updating step improves accuracy of likelihood approximation when the data size $H=8$. With the help of MALA, $\log (\theta_3)$ converges close to the true value when $H=8,16$. Additionally, for $\log (\theta_1)$ and $\log (\theta_2)$ under all data sizes, based on the likelihood approximated by the Bayesian updating LNA metamodel, MALA performs better than M-H in terms of both RMSEs and their confidence half-widths, meaning that MALA converges faster than M-H; while for $\log (\theta_3)$ and $\log (\sigma)$, the performance of the two algorithms is similar. This is because both two algorithms have converged before $T_0=10000$.


\begin{table*}[th]
    \caption{The RMSEs between the estimated and the true log-parameters (with 95\% CIs).}
    \scriptsize
    \centering
    \label{table:rmse}
    \begin{tabular}{c|c|c|c|c|c|c|c|c|c}
        \toprule
        & \multicolumn{3}{c|}{MALA with Bayesian updating LNA} & \multicolumn{3}{c|}{M-H with Bayesian updating LNA} & \multicolumn{3}{c}{{MALA with original LNA}} \\ \midrule
        Data & $H=4$ & $H=8$ & $H=16$ & $H=4$ & $H=8$ & $H=16$ & {$H=4$} & {$H=8$} & {$H=16$} \\ 
        size & $(\Delta t = 20)$ & $(\Delta t = 10)$ & $(\Delta t = 5)$ & $(\Delta t = 20)$ & $(\Delta t = 10)$ & $(\Delta t = 5)$ & {$(\Delta t = 20)$} & {$(\Delta t = 10)$} & {$(\Delta t = 5)$} \\ \midrule
        $\log (\theta_1)$ & $1.79 \pm 0.73$ & $1.27 \pm 0.92$ & $0.48 \pm 0.08$ & $2.56 \pm 1.09$ & $1.80 \pm 1.07$ & $1.48 \pm 1.37$ & {$1.82 \pm 0.74$} & {$1.72 \pm 0.81$} & {$1.41 \pm 0.94$} \\ \midrule
        $\log (\theta_2)$ & $2.76 \pm 1.37$ & $2.12 \pm 1.31$ & $1.32 \pm 0.58$ & $3.29 \pm 1.82$ & $2.87 \pm 1.49$ & $2.50 \pm 1.61$ & {$2.80 \pm 1.38$} & {$2.73 \pm 1.36$} & {$2.63 \pm 1.19$} \\ \midrule
        $\log (\theta_3)$ & $1.73 \pm 2.02$ & $0.25 \pm 0.03$ & $0.28 \pm 0.03$ & $1.66 \pm 2.12$ & $1.15 \pm 2.02$ & $0.24 \pm 0.03$ & {$1.71 \pm 2.02$} & {$0.69 \pm 0.83$} & {$0.17 \pm 0.04$} \\ \midrule
        $\log (\sigma)$ & $1.81 \pm 0.93$ & $1.02 \pm 0.56$ & $0.92 \pm 0.59$ & $1.59 \pm 0.70$ & $0.98 \pm 0.43$ & $0.80 \pm 0.29$ & {$1.76 \pm 0.93$} & {$1.30 \pm 0.62$} & {$1.05 \pm 0.73$} \\ \bottomrule
    \end{tabular}
\end{table*}

\section{Conclusion}
\label{sec: conclusion}

Bayesian inference on partially observed SRN plays a critical role for multi-scale bioprocess mechanism learning. To tackle the critical challenges of biomanufacturing processes, including high complexity, high inherent stochasticity, and very limited and sparse observations on partially observed state with measurement errors, we propose an interpretable Bayesian updating LNA metamodel to approximate the likelihood of heterogeneous online and offline measures, accounting for the structure information of the enzymatic SRN mechanistic model. Then, we develop a MALA sampling algorithm that utilizes the information from the derived likelihood and more efficiently generates posterior samples. The empirical study shows that our proposed LNA assisted Bayesian inference approach has a promising performance, demonstrating its potential to benefit bioprocess mechanisms online learning and digital twin development.

\section*{Acknowledgements}

The authors thank Prof. Cheng Li from the Department of Statistics and Data Science, National University of Singapore for providing the discussion of Metropolis-adjusted Langevin algorithm (MALA).


\footnotesize

\bibliographystyle{unsrt}

\bibliography{demobib}

\begin{thebibliography}{10}

\bibitem{hillson2019building}
Nathan Hillson, Mark Caddick, Yizhi Cai, Jose~A Carrasco, Matthew~Wook Chang, Natalie~C Curach, David~J Bell, Rosalind Le~Feuvre, Douglas~C Friedman, Xiongfei Fu, et~al.
\newblock Building a global alliance of biofoundries.
\newblock {\em Nature Communications}, 10(1):2040, 2019.

\bibitem{gillespie2000chemical}
Daniel~T Gillespie.
\newblock The chemical langevin equation.
\newblock {\em The Journal of Chemical Physics}, 113(1):297--306, 2000.

\bibitem{xie2022sequential}
Wei Xie, Keqi Wang, Hua Zheng, and Ben Feng.
\newblock Sequential importance sampling for hybrid model bayesian inference to support bioprocess mechanism learning and robust control.
\newblock In {\em 2022 Winter Simulation Conference (WSC)}, pages 2282--2293. IEEE, 2022.

\bibitem{archambeau2007gaussian}
Cedric Archambeau, Dan Cornford, Manfred Opper, and John Shawe-Taylor.
\newblock Gaussian process approximations of stochastic differential equations.
\newblock In {\em Gaussian Processes in Practice}, pages 1--16. PMLR, 2007.

\bibitem{garcia2017nonparametric}
Constantino~A Garcia, Abraham Otero, Paulo Felix, Jesus Presedo, and David~G Marquez.
\newblock Nonparametric estimation of stochastic differential equations with sparse gaussian processes.
\newblock {\em Physical Review E}, 96(2):022104, 2017.

\bibitem{yang2021inference}
Shihao Yang, Samuel~WK Wong, and SC~Kou.
\newblock Inference of dynamic systems from noisy and sparse data via manifold-constrained gaussian processes.
\newblock {\em Proceedings of the National Academy of Sciences}, 118(15):e2020397118, 2021.

\bibitem{gillespie1992rigorous}
Daniel~T Gillespie.
\newblock A rigorous derivation of the chemical master equation.
\newblock {\em Physica A: Statistical Mechanics and Its Applications}, 188(1-3):404--425, 1992.

\bibitem{ferm2008hierarchy}
Lars Ferm, Per L{\"o}tstedt, and Andreas Hellander.
\newblock A hierarchy of approximations of the master equation scaled by a size parameter.
\newblock {\em Journal of Scientific Computing}, 34(2):127--151, 2008.

\bibitem{ruttor2009efficient}
Andreas Ruttor and Manfred Opper.
\newblock Efficient statistical inference for stochastic reaction processes.
\newblock {\em Physical Review Letters}, 103(23):230601, 2009.

\bibitem{fearnhead2014inference}
Paul Fearnhead, Vasilieos Giagos, and Chris Sherlock.
\newblock Inference for reaction networks using the linear noise approximation.
\newblock {\em Biometrics}, 70(2):457--466, 2014.

\bibitem{chewi2021optimal}
Sinho Chewi, Chen Lu, Kwangjun Ahn, Xiang Cheng, Thibaut Le~Gouic, and Philippe Rigollet.
\newblock Optimal dimension dependence of the metropolis-adjusted langevin algorithm.
\newblock In {\em Conference on Learning Theory}, pages 1260--1300. PMLR, 2021.

\bibitem{anderson2011continuous}
David~F Anderson and Thomas~G Kurtz.
\newblock Continuous time markov chain models for chemical reaction networks.
\newblock In {\em Design and Analysis of Biomolecular Circuits: Engineering Approaches to Systems and Synthetic Biology}, pages 3--42. Springer, 2011.

\bibitem{kloeden1992numerical}
P.E. Kloeden and E.~Platen.
\newblock {\em Numerical Solution of Stochastic Differential Equations}.
\newblock Applications of Mathematics : Stochastic Modelling and Applied Probability. Springer, 1992.

\bibitem{prague2013nimrod}
M{\'e}Lanie Prague, Daniel Commenges, J{\'e}R{\'e}Mie Guedj, Julia Drylewicz, and Rodolphe Thi{\'e}Baut.
\newblock Nimrod: A program for inference via a normal approximation of the posterior in models with random effects based on ordinary differential equations.
\newblock {\em Computer Methods and Programs in Biomedicine}, 111(2):447--458, 2013.

\bibitem{rao2003stochastic}
Christopher~V Rao and Adam~P Arkin.
\newblock Stochastic chemical kinetics and the quasi-steady-state assumption: Application to the gillespie algorithm.
\newblock {\em The Journal of Chemical Physics}, 118(11):4999--5010, 2003.

\bibitem{gillespie1977exact}
Daniel~T Gillespie.
\newblock Exact stochastic simulation of coupled chemical reactions.
\newblock {\em The Journal of Physical Chemistry}, 81(25):2340--2361, 1977.

\end{thebibliography}

\end{document}